\documentclass[10pt,twocolumn,letterpaper]{article}

\usepackage[pagenumbers]{cvpr} 

\usepackage{multirow}
\usepackage{makecell}
\usepackage{array}
\usepackage{graphicx}
\usepackage{amsmath}
\usepackage{amssymb}
\usepackage{booktabs}
\usepackage{tabularx}

\definecolor{cvprblue}{rgb}{0.21,0.49,0.74}
\usepackage[pagebackref,breaklinks,colorlinks,allcolors=cvprblue]{hyperref}

\def\paperID{*****} 
\def\confName{CVPR}
\def\confYear{2026}

\title{On the Efficacy of Self-Supervised Point Cloud Encoders\\for Efficient 3D Large Language Models}

\author{
Yao Zheng\textsuperscript{1}\quad
Tian Zhang\textsuperscript{1}\\
\textsuperscript{1}Beijing University of Posts and Telecommunications\\
{\tt\small \{3448961153, ztian\}@bupt.edu.cn}
}

\begin{document}
\maketitle
\begin{abstract}
Large 3D point cloud-language models (3D-LLMs) have demonstrated remarkable capabilities in 3D understanding by integrating point cloud encoders with large language models. However, existing 3D-LLMs rely on multi-modal pre-trained encoders (\eg, ULIP-2) that require expensive image-text-point cloud alignment with 8$\times$ A100-scale computing, raising the barrier for research and deployment. In this work, we conduct a systematic empirical study to investigate whether low-cost self-supervised point cloud encoders---specifically PCP-MAE and Point-MAE---can serve as effective alternatives. Using MiniGPT-3D as the testbed, we pre-train and evaluate 7 encoder initialization / pre-training configurations (1 multi-modal baseline + 5 self-supervised pre-training + 1 random initialization), evaluated under 2 fine-tuning strategies (frozen vs.\ unfrozen) for a total of 12 experimental groups, spanning 2 architectures (MaskTransformer vs.\ PointTransformer), 3 pre-training objectives (PCP-MAE, Point-MAE, random initialization), and 2 pre-training datasets (Objaverse 660K vs.\ ShapeNet55-34 $\sim$50K). Our experiments reveal three principal findings: \textbf{(1) The four-stage MiniGPT-3D training pipeline is remarkably capable of training a 3D encoder from random initialization}---a randomly initialized encoder trained end-to-end achieves 52.50\% open-vocabulary accuracy and 44.45 captioning score, approaching the best pre-trained variants; \textbf{(2) Architecture and pre-training objective exhibit a strong crossover interaction}---PCP-MAE with MaskTransformer achieves 59.00\% accuracy (the best self-supervised result), while Point-MAE with MaskTransformer drops to 46.50\%, and the pattern inverts for PointTransformer; \textbf{(3) Closed-set ModelNet40 classification remains a fundamental weakness of all purely geometric encoders}, achieving only $\sim$13--18\% accuracy compared to $\sim$62\% for the multi-modal baseline, even after end-to-end fine-tuning. Our findings provide actionable guidelines for building cost-effective 3D-LLMs and reveal interaction patterns between self-supervised pre-training objectives and downstream encoder architectures.


\end{abstract}

\section{Introduction}
\label{sec:intro}

Large Language Models (LLMs) have driven remarkable advancements across multiple domains, benefiting from their extensive world knowledge and instruction-following capabilities \cite{touvron2023llama, openai2023gpt4}. Building upon LLMs, large 2D vision-language models (2D-LLMs) successfully bridge visual and textual modalities through learned projectors, enabling comprehensive visual understanding \cite{li2023blip2, dai2023instructblip}. Inspired by these successes, large 3D point cloud-language models (3D-LLMs) have emerged to equip LLMs with the ability to perceive and reason about 3D geometry \cite{qi2024shapellm, xu2023pointllm, tang2024minigpt3d}.

A critical component in 3D-LLMs is the point cloud encoder, which transforms raw 3D point clouds into feature representations. State-of-the-art 3D-LLMs such as PointLLM \cite{xu2023pointllm} and ShapeLLM \cite{qi2024shapellm} employ point cloud encoders pre-trained with multi-modal objectives (\eg, ULIP-2 \cite{xue2024ulip2}) that align point clouds with images and text. While effective, such multi-modal pre-training is computationally expensive and data-intensive, requiring carefully curated tri-modal datasets of $\sim$1.2TB. MiniGPT-3D \cite{tang2024minigpt3d} proposed an efficient alternative by leveraging 2D-LLM priors to reduce the cost of vision-language alignment, but it still depends on the ULIP-2 pre-trained Point-BERT \cite{yu2022pointbert} encoder.

In parallel, self-supervised learning methods for 3D point clouds---such as Point-MAE \cite{pang2022pointmae} and PCP-MAE \cite{zhang2024pcpmae}---have demonstrated strong representation learning capabilities through masked autoencoding and geometric reconstruction, requiring only unlabeled point cloud data. However, their effectiveness as encoder backbones in 3D-LLM pipelines remains unexplored.

In this work, we conduct a systematic empirical study using MiniGPT-3D \cite{tang2024minigpt3d} as our testbed, pre-training and evaluating 7 encoder initialization / pre-training configurations (1 multi-modal baseline + 5 self-supervised pre-training + 1 random initialization), evaluated under 2 fine-tuning strategies (frozen vs.\ unfrozen) for a total of 12 experimental groups. Our study is guided by three questions: \textbf{(Q1)} Can self-supervised encoders match multi-modal encoders when the downstream pipeline is allowed to fine-tune them? \textbf{(Q2)} How do architecture choice (MaskTransformer vs.\ PointTransformer) and pre-training objective (PCP-MAE vs.\ Point-MAE) interact? \textbf{(Q3)} What is the marginal benefit of self-supervised pre-training over random initialization in this setup?

\textbf{Our main contributions are:}

\begin{enumerate}
\item \textbf{We discover that the MiniGPT-3D pipeline can train a 3D encoder from random initialization to nearly match pre-trained variants.} A randomly initialized encoder, trained end-to-end through the four stages, achieves 52.50\% open-vocabulary classification accuracy---matching the PCP-MAE pre-trained V2 (52.50\%) and closely approaching the best self-supervised variant (59.00\%). This suggests that the strong 2D-LLM priors and cascaded training largely compensate for the absence of 3D pre-training.

\item \textbf{We reveal a previously undocumented crossover interaction between architecture and pre-training objective.} PCP-MAE center prediction synergizes with MaskTransformer's cross-attention (59.00\% accuracy), while pure masked reconstruction (Point-MAE) performs best with PointTransformer's self-attention (55.50\%). The optimal architecture flips depending on the pre-training objective---a finding with significant implications for future 3D representation learning.

\item \textbf{We characterize the hard ceiling of geometric pre-training for closed-set classification.} All self-supervised variants, regardless of architecture, objective, data scale, or fine-tuning strategy, plateau at 13--18\% on ModelNet40 (vs.\ 62\% baseline), confirming that semantic category boundaries require multi-modal signals that geometric reconstruction alone cannot provide.
\end{enumerate}

\section{Related Work}

\subsection{Large 3D Point Cloud-Language Models}

Large 3D point cloud-language models (3D-LLMs) integrate point cloud understanding into LLMs, enabling reasoning and conversation about 3D objects \cite{qi2024shapellm, xu2023pointllm, tang2024minigpt3d, hong2023three, chen2023ll3da}. Early work by 3D-LLM \cite{hong2023three} renders 3D objects into multi-view 2D images and leverages 2D-LLM priors as an intermediary between LLMs and 3D data. Subsequent approaches \cite{xu2023pointllm, chen2023ll3da} encode point clouds directly using 3D encoders and align them with LLMs through trainable projectors.

PointLLM \cite{xu2023pointllm} and ShapeLLM \cite{qi2024shapellm} demonstrate that fully fine-tuning 3D-LLMs with large LLM backbones (7B--13B parameters) yields strong performance but requires substantial computational resources. MiniGPT-3D \cite{tang2024minigpt3d} addresses efficiency by proposing a four-stage cascaded training strategy that leverages 2D-LLM priors, achieving strong performance with only 26.8 hours of training on a single RTX 3090 GPU. Despite this efficiency gain, MiniGPT-3D still relies on the ULIP-2 \cite{xue2024ulip2} pre-trained Point-BERT \cite{yu2022pointbert} encoder, which itself requires expensive multi-modal pre-training. All existing 3D-LLMs default to multi-modal pre-trained encoders, leaving the feasibility of cheaper geometric-only encoders largely unexplored.

\subsection{Self-Supervised Point Cloud Representation Learning}

Self-supervised learning for 3D point clouds has advanced rapidly, with masked autoencoding emerging as a dominant paradigm. Point-BERT \cite{yu2022pointbert} adapts BERT-style masked modeling to point clouds using a discrete tokenizer. Point-MAE \cite{pang2022pointmae} simplifies this by directly reconstructing masked point coordinates through an asymmetric encoder-decoder architecture. PCP-MAE \cite{zhang2024pcpmae} extends Point-MAE with an auxiliary center prediction task that predicts the centroid of masked point patches.

Two encoder architectures are commonly used: \textbf{PointTransformer} (self-attention over all patches, used in Point-BERT/Point-MAE) and \textbf{MaskTransformer} (cross-attention between visible and masked patches, used in PCP-MAE). Their compatibility with downstream 3D-LLM pipelines has not been systematically compared.

\subsection{Efficient Vision-Language Alignment}

Parameter-efficient fine-tuning (PEFT) techniques such as LoRA \cite{hu2022lora} have been widely adopted in vision-language models. TinyGPT-V \cite{yuan2023tinygptv} and MobileVLM \cite{chu2023mobilevlm} demonstrate that efficient 2D-LLMs can be built by combining pre-trained vision modules with LoRA-adapted LLMs. MiniGPT-3D \cite{tang2024minigpt3d} extends this to 3D by employing LoRA and Norm fine-tuning, achieving up to 260$\times$ fewer trainable parameters. Our work extends this efficiency research upstream, exploring whether the encoder itself can be made cheaper to pre-train.

\section{Methodology}

\subsection{Preliminaries: MiniGPT-3D}

MiniGPT-3D \cite{tang2024minigpt3d} aligns 3D point clouds with LLMs using 2D priors. Its architecture consists of: a point cloud encoder, a point cloud projection layer (MLP), a Q-Former \cite{li2023blip2}, a mixture of query experts (MQE), a modality projector, and a large language model (Phi-2, 2.7B). The training proceeds in four stages: (I) training the point cloud projection layer; (II) fine-tuning the Q-Former, modality projector, and LLM; (III) fine-tuning on complex instructions and detailed captions; (IV) training the MQE module.

In the original MiniGPT-3D, the point cloud encoder is Point-BERT pre-trained with ULIP-2 on Objaverse, kept frozen throughout all stages. We replace this encoder with self-supervised alternatives and examine both frozen and unfrozen configurations.

\begin{figure*}[t]
  \centering
  \includegraphics[width=\textwidth]{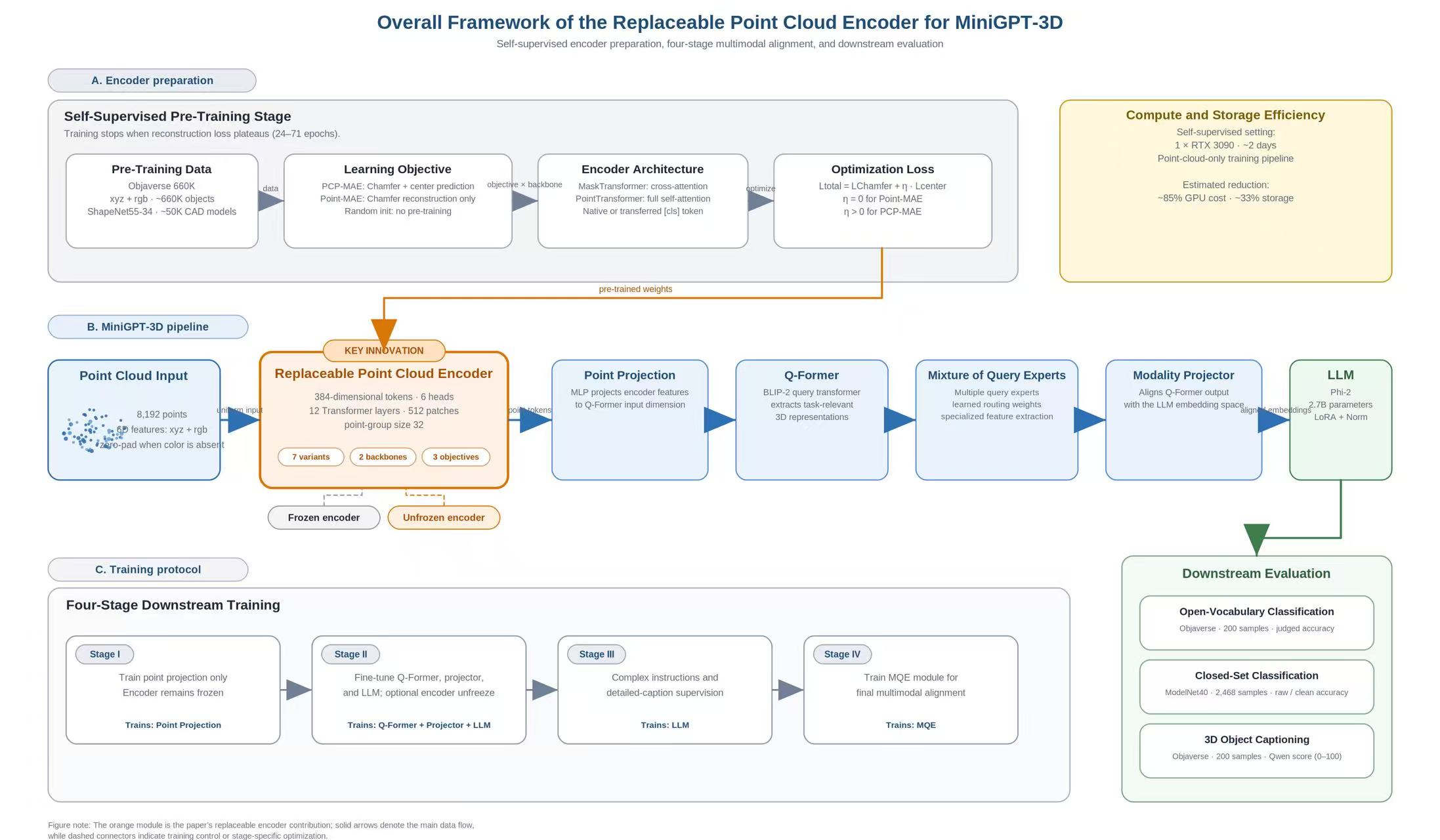}
  \caption{\textbf{Complete MiniGPT-3D Pipeline Architecture.} The framework consists of: (1) Input layer accepting 8,192 points $\times$ 6D features; (2) Replaceable Point Cloud Encoder (innovation hub) with 7 encoder variants spanning 2 architectures (MaskTransformer vs.\ PointTransformer), 3 pre-training objectives (PCP-MAE/Point-MAE/Random Init), and 2 datasets (Objaverse 660K/ShapeNet55-34 50K); (3) Downstream pipeline: MLP Projection $\rightarrow$ Q-Former $\rightarrow$ MQE $\rightarrow$ Modality Projector $\rightarrow$ LLM (Phi-2 2.7B); (4) Four training stages with frozen/unfrozen strategies; (5) Three evaluation tasks: Open-Vocab Classification, ModelNet40 Closed-Set, and 3D Captioning; (6) Five key findings with supporting data; (7) Cost comparison: 8$\times$A100 (ULIP-2) vs.\ 1$\times$RTX 3090 (self-supervised).}
  \label{fig:framework}
\end{figure*}

\subsection{Self-Supervised Encoder Variants}

We pre-train 7 encoder initialization / pre-training configurations (1 multi-modal baseline + 5 self-supervised pre-training + 1 random initialization), evaluated under both frozen and unfrozen strategies for a total of 12 experimental groups. All accept 8,192 points with 6-dimensional features (xyz + rgb; RGB zero-padded for datasets lacking color). Pre-training is halted when the reconstruction loss enters a stable plateau (24--71 epochs).

The pre-training loss is $\mathcal{L}_{total} = \mathcal{L}_{chamfer} + \eta \cdot \mathcal{L}_{center}$, where $\eta$ controls the center prediction weight. For Point-MAE, $\eta = 0$.

\begin{table*}[t]
  \centering
  \small
  \caption{\textbf{Encoder variants evaluated in this study.} The ``Cls Token'' column reflects a fundamental architectural distinction between the two encoder backbones. \textbf{PointTransformer} (used in V2, Point-MAE, Random unfrozen, and the Baseline) is designed for downstream fine-tuning: it prepends a learnable [cls] token to the sequence of all patch tokens and applies uniform self-attention over the full sequence, allowing the [cls] token to aggregate global information---analogous to the [cls] token in ViT/BERT. \textbf{MaskTransformer} (used in V1 hybrid, Mask Point-MAE, ShapeNet55-34) is designed purely for masked autoencoding: it splits tokens into visible and masked groups, processes only visible patches through self-attention, and lets masked tokens attend to visible ones via cross-attention. There is no [cls] token because the pre-training objective is local patch reconstruction, not global classification. However, because MiniGPT-3D's encoder interface (the \texttt{PointTransformer} class in \texttt{point\_encoder.py}) always expects \texttt{cls\_token} and \texttt{cls\_pos} parameters, MaskTransformer variants that lack these native parameters must obtain them by copying from the Baseline checkpoint (V1 hybrid, Mask Point-MAE) or by random initialization (ShapeNet55-34, Random unfrozen). This grafting is necessary for compatibility but provides no meaningful global representation, as confirmed by the near-orthogonal cls features reported in Section~4.3.1.}
  \label{tab:variants}
  \begin{tabular}{lllll}
    \toprule
    \textbf{Variant} & \textbf{Objective} & \textbf{Architecture} & \textbf{Pre-training Data} & \textbf{Cls Token} \\
    \midrule
    Baseline & ULIP-2 (multi-modal) & PointTransformer & Objaverse 660K & Native \\
    V1 hybrid & PCP-MAE & MaskTransformer & Objaverse 660K & Copied from Baseline \\
    V1 hybrid unfrozen & PCP-MAE & MaskTransformer & Objaverse 660K & Copied from Baseline \\
    V2 & PCP-MAE & PointTransformer & Objaverse 660K & Native \\
    V2 unfrozen & PCP-MAE & PointTransformer & Objaverse 660K & Native \\
    Point-MAE & Point-MAE ($\eta=0$) & PointTransformer & Objaverse 660K & Native \\
    Point-MAE unfrozen & Point-MAE ($\eta=0$) & PointTransformer & Objaverse 660K & Native \\
    Mask Point-MAE & Point-MAE ($\eta=0$) & MaskTransformer & Objaverse 660K & Copied from Baseline \\
    Mask Point-MAE unfrozen & Point-MAE ($\eta=0$) & MaskTransformer & Objaverse 660K & Copied from Baseline \\
    ShapeNet55-34 & PCP-MAE & MaskTransformer & ShapeNet55-34 ($\sim$50K) & Random init \\
    ShapeNet55-34 unfrozen & PCP-MAE & MaskTransformer & ShapeNet55-34 ($\sim$50K) & Random init \\
    Random unfrozen & None (random init) & PointTransformer & -- & Random init \\
    \bottomrule
  \end{tabular}
\end{table*}

\subsubsection{V1 Hybrid: PCP-MAE + MaskTransformer}

Uses the original PCP-MAE MaskTransformer architecture with cross-attention between visible and masked tokens. Lacks a native [cls] token; we copy these parameters from the Baseline. Pre-trained on Objaverse 660K for 24 epochs (Chamfer loss: 1095$\rightarrow$0.57; center loss: 24.35$\rightarrow$1.41). We refer to this variant as ``hybrid'' because it combines the MaskTransformer backbone (from PCP-MAE) with the cls token parameters copied from the Baseline.

\subsubsection{V2: PCP-MAE + PointTransformer}

Replaces MaskTransformer with PointTransformer (self-attention + native cls token), matching the Baseline architecture. Pre-trained on Objaverse 660K for 32 epochs (Chamfer loss: 815$\rightarrow$0.82; center loss: 34.5$\rightarrow$0.046).

\subsubsection{Point-MAE: Ablation of Center Prediction ($\eta=0$)}

Identical to V2 but with $\eta=0$, reducing the objective to pure Point-MAE reconstruction. Pre-trained on Objaverse 660K for 35 epochs.

\subsubsection{Mask Point-MAE: Pure Reconstruction with MaskTransformer}

Identical to V1 hybrid but with $\eta=0$. Pre-trained on Objaverse 660K for 31 epochs.

\subsubsection{ShapeNet55-34: Cross-Dataset Comparison}

PCP-MAE MaskTransformer pre-trained on ShapeNet55-34 ($\sim$50K samples, 13$\times$ smaller than Objaverse, no color). Pre-trained for 71 epochs.

\subsubsection{Random Unfrozen: No Pre-training Baseline}

A PointTransformer with randomly initialized weights, serving as the lower bound for end-to-end training.

\subsection{Training and Evaluation Protocol}

We first fully reproduce the Baseline (MiniGPT-3D with ULIP-2 pre-trained Point-BERT encoder) by re-running the complete four-stage training pipeline, obtaining results consistent with the official weights evaluation (Table~A1 in Appendix). All subsequent variant experiments are compared against this reproduced Baseline, ensuring that any performance differences stem solely from the encoder itself rather than the training procedure.

All variants undergo the identical MiniGPT-3D four-stage pipeline on a single RTX 3090. \textbf{Frozen} variants keep the encoder fixed; \textbf{unfrozen} variants allow encoder weight updates during stages II--IV.

Three evaluation tasks follow the standard MiniGPT-3D protocol:

\begin{itemize}
\item \textbf{Open-Vocabulary Classification (Objaverse, 200 samples):} Generate text descriptions and evaluate accuracy via Qwen-Flash API. Two prompts: ``What is this?'' (Instruction) and ``This is an object of'' (Completion).
\item \textbf{Closed-Set Classification (ModelNet40, 2,468 samples, 40 classes):} Generate responses and map to 40 classes via Qwen-Flash API evaluation. Report raw accuracy and clean accuracy (excluding invalid responses, defined as responses that cannot be mapped to any of the 40 predefined classes).
\item \textbf{3D Object Captioning (Objaverse, 200 samples):} ``Caption this 3D model in detail.'' Captions scored by Qwen-Flash API (0--100).
\end{itemize}

All subjective evaluations use the Alibaba Cloud Bailian Qwen-Flash API with consistent parameters (temperature, max tokens, prompt templates) across all variants. The LLM judge evaluates responses against the ground truth label to determine correctness for classification tasks, and scores caption quality holistically on a 0--100 scale. All variants use identical evaluation prompts, API parameters, and scoring criteria, ensuring that relative comparisons between variants are valid even though absolute scores differ from GPT-4-based evaluations in prior work.

\subsection{Encoder Quality Diagnosis}

Prior to MiniGPT-3D training, we analyze each encoder's feature space on 2,468 ModelNet40 samples:
\begin{enumerate}
\item \textbf{Cross-encoder cosine similarity} at multiple representation levels (cls, global, router)
\item \textbf{kNN classification} accuracy on ModelNet40
\item \textbf{Intra-class vs.\ inter-class cosine similarity} to quantify class separability
\end{enumerate}

\section{Experiments}
\label{sec:experiments}

\subsection{Main Results}
\label{sec:main_results}

\begin{figure*}[t]
  \centering
  \includegraphics[width=\textwidth]{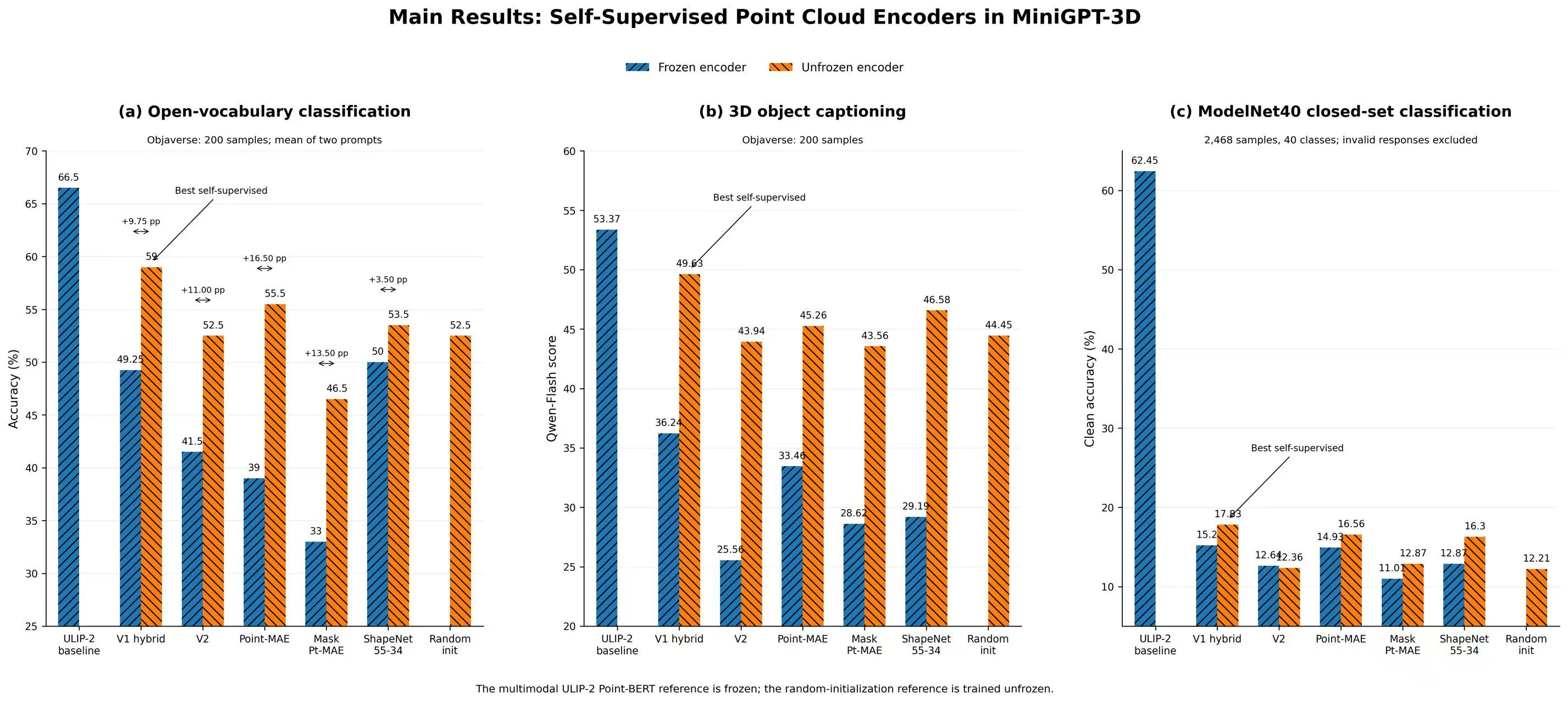}
  \caption{\textbf{Main Experimental Results Across 12 Variant Groups.} Three sub-figures present: (a) Open-Vocabulary Classification accuracy, (b) 3D Captioning scores, and (c) ModelNet40 Closed-Set accuracy. Each of the 12 variant nodes is annotated with raw metrics from Tables~2--4. PERFORMANCE\_IN edges link each variant to each sub-figure with value and group tags. Frozen/Unfrozen grouping enables side-by-side comparison. Delta analysis nodes quantify frozen$\rightarrow$unfrozen gains (avg +10.9pp OV, +15.2 caption, +1.9 MN40). Key insights: universal unfreezing benefit, random init competitiveness, ModelNet40 ceiling, and data quality vs.\ scale trade-off.}
  \label{fig:main_results}
\end{figure*}

\subsubsection{Open-Vocabulary 3D Object Classification}

\begin{table*}[t]
  \centering
  \small
  \caption{Open-vocabulary classification accuracy (\%) on Objaverse. I = Instruction; C = Completion. Variants grouped by frozen / unfrozen strategy. \textbf{Best self-supervised result in bold.}}
  \label{tab:open_vocab}
  \begin{tabular}{lccc}
    \toprule
    \textbf{Encoder} & \textbf{Prompt 0 (I)} & \textbf{Prompt 1 (C)} & \textbf{Avg} \\
    \midrule
    \textbf{Frozen} & & & \\
    Baseline (ULIP-2 Point-BERT) & 67.00\% & 66.00\% & 66.50\% \\
    V1 hybrid (PCP-MAE, MaskTransformer) & 47.50\% & 51.00\% & 49.25\% \\
    V2 (PCP-MAE, PointTransformer) & 39.00\% & 44.00\% & 41.50\% \\
    Point-MAE (Point-MAE, PointTransformer) & 36.00\% & 42.00\% & 39.00\% \\
    Mask Point-MAE (Point-MAE, MaskTransformer) & 34.00\% & 32.00\% & 33.00\% \\
    ShapeNet55-34 (PCP-MAE, MaskTransformer) & 49.50\% & 50.50\% & 50.00\% \\
    \textbf{Unfrozen} & & & \\
    V1 hybrid (PCP-MAE, MaskTransformer) & \textbf{59.00\%} & \textbf{59.00\%} & \textbf{59.00\%} \\
    V2 (PCP-MAE, PointTransformer) & 52.50\% & 52.50\% & 52.50\% \\
    Point-MAE (Point-MAE, PointTransformer) & 54.50\% & 56.50\% & 55.50\% \\
    Mask Point-MAE (Point-MAE, MaskTransformer) & 46.50\% & 46.50\% & 46.50\% \\
    ShapeNet55-34 (PCP-MAE, MaskTransformer) & 54.50\% & 52.50\% & 53.50\% \\
    Random (PointTransformer) & 52.50\% & 52.50\% & 52.50\% \\
    \bottomrule
  \end{tabular}
\end{table*}

\textbf{Finding 1: The frozen $\rightarrow$ unfrozen gap is large and universal.} Unfreezing consistently boosts all variants: V1 hybrid +9.75pp, V2 +11.00pp, Point-MAE +16.50pp, Mask Point-MAE +13.50pp, ShapeNet55-34 +3.50pp.

\textbf{Finding 2: Random initialization matches two pre-trained variants.} The Random unfrozen encoder (52.50\%) exactly matches V2 unfrozen (52.50\%), and narrowly trails V1 hybrid unfrozen (59.00\%) by only 6.50pp. This is a striking result: \textbf{the MiniGPT-3D pipeline's four-stage training, combined with strong 2D-LLM priors, can train a 3D encoder from random initialization to match a PCP-MAE pre-trained PointTransformer, and reach 88\% of the best self-supervised variant's performance.}

\textbf{Finding 3: A strong crossover interaction between architecture and pre-training objective emerges.} In the unfrozen setting:
\begin{itemize}
\item PCP-MAE (with center prediction): MaskTransformer (59.00\%) $\gg$ PointTransformer (52.50\%)
\item Point-MAE (pure reconstruction): PointTransformer (55.50\%) $\gg$ MaskTransformer (46.50\%)
\end{itemize}

This complete inversion reveals that the center prediction task synergizes with MaskTransformer's cross-attention mechanism, while pure masked reconstruction is better served by PointTransformer's uniform self-attention. We analyze this phenomenon in detail in Section~4.2.2.

\textbf{Finding 4: ShapeNet55-34 (50K samples) is competitive with Objaverse (660K).} ShapeNet55-34 unfrozen (53.50\%) exceeds V2 unfrozen (52.50\%) and approaches V1 hybrid unfrozen (59.00\%), despite a 13$\times$ disadvantage in data volume.

\subsubsection{Closed-Set Classification on ModelNet40}

\begin{table*}[t]
  \centering
  \small
  \caption{Closed-set classification accuracy (\%) on ModelNet40. ``acc'' = raw accuracy; ``clean'' = accuracy excluding invalid responses (responses that cannot be mapped to any of the 40 predefined classes). Variants grouped by frozen / unfrozen strategy. \textbf{Best self-supervised clean accuracy in bold.}}
  \label{tab:modelnet40}
  \begin{tabular}{lcc}
    \toprule
    \textbf{Encoder} & \textbf{Prompt 0 (acc / clean)} & \textbf{Prompt 1 (acc / clean)} \\
    \midrule
    \textbf{Frozen} & & \\
    Baseline (ULIP-2 Point-BERT) & 61.35\% / 63.83\% & 59.08\% / 61.06\% \\
    V1 hybrid (PCP-MAE, MaskTransformer) & 10.98\% / 14.62\% & 13.17\% / 15.77\% \\
    V2 (PCP-MAE, PointTransformer) & 10.21\% / 12.72\% & 9.93\% / 12.55\% \\
    Point-MAE (Point-MAE, PointTransformer) & 11.10\% / 14.14\% & 12.20\% / 15.71\% \\
    Mask Point-MAE (Point-MAE, MaskTransformer) & 8.43\% / 9.96\% & 10.45\% / 12.06\% \\
    ShapeNet55-34 (PCP-MAE, MaskTransformer) & 10.45\% / 12.84\% & 10.70\% / 12.89\% \\
    \textbf{Unfrozen} & & \\
    V1 hybrid (PCP-MAE, MaskTransformer) & 13.29\% / \textbf{18.41\%} & 13.41\% / \textbf{17.25\%} \\
    V2 (PCP-MAE, PointTransformer) & 10.05\% / 12.50\% & 9.85\% / 12.22\% \\
    Point-MAE (Point-MAE, PointTransformer) & 13.98\% / 17.11\% & 13.70\% / 16.01\% \\
    Mask Point-MAE (Point-MAE, MaskTransformer) & 10.13\% / 13.10\% & 10.25\% / 12.63\% \\
    ShapeNet55-34 (PCP-MAE, MaskTransformer) & 12.16\% / 16.94\% & 12.03\% / 15.65\% \\
    Random (PointTransformer) & 10.45\% / 13.67\% & 9.16\% / 10.74\% \\
    \bottomrule
  \end{tabular}
\end{table*}

\textbf{Finding: A hard ceiling at $\sim$18\%.} All self-supervised variants cluster tightly between 10\% and 18\% clean accuracy, regardless of architecture, pre-training objective, dataset scale, or fine-tuning strategy. Even unfreezing provides only marginal gains (\eg, V1 hybrid 14.62\% $\rightarrow$ 18.41\%). The Baseline achieves 61--64\%, a gap of 43--47 percentage points that no geometric pre-training method can bridge.

\textbf{Two complementary perspectives explain this ceiling:}

\textit{Invalid response perspective:} The self-supervised variants produce substantially higher invalid response rates (20--30\%, i.e., responses that do not map to any of the 40 classes) compared to the Baseline ($\sim$3\%). This indicates that the model cannot extract clear semantic category boundaries from purely geometric features, causing the LLM to produce outputs outside the predefined class set.

\textit{Feature margin perspective:} As analyzed in Section~4.3.3, the self-supervised encoders have $\sim$23--26\% lower class separability margin (0.087--0.089 vs.\ 0.116). The $\sim$26\% reduction in feature-level margin cascades through the downstream alignment layers, ultimately resulting in a $\sim$70\% absolute accuracy drop on ModelNet40 (62\% $\rightarrow$ 18\%). This establishes a clear causal chain from feature representation to closed-set classification performance.

\textbf{Interpretation: Precise categorization across 40 predefined classes requires semantic boundaries that only multi-modal (image-text-point cloud) alignment provides.}

\subsubsection{3D Object Captioning}

\begin{table}[t]
  \centering
  \small
  \caption{3D object captioning scores (0--100) on Objaverse, evaluated by Qwen-Flash API. Variants grouped by frozen / unfrozen strategy. \textbf{Best self-supervised result in bold.}}
  \label{tab:captioning}
  \resizebox{\columnwidth}{!}{%
  \begin{tabular}{lc}
    \toprule
    \textbf{Encoder} & \textbf{Average Score} \\
    \midrule
    \textbf{Frozen} & \\
    Baseline (ULIP-2 Point-BERT) & 53.37 \\
    V1 hybrid (PCP-MAE, MaskTransformer) & 36.24 \\
    V2 (PCP-MAE, PointTransformer) & 25.56 \\
    Point-MAE (Point-MAE, PointTransformer) & 33.46 \\
    Mask Point-MAE (Point-MAE, MaskTransformer) & 28.62 \\
    ShapeNet55-34 (PCP-MAE, MaskTransformer) & 29.19 \\
    \textbf{Unfrozen} & \\
    \textbf{V1 hybrid (PCP-MAE, MaskTransformer)} & \textbf{49.63} \\
    V2 (PCP-MAE, PointTransformer) & 43.94 \\
    Point-MAE (Point-MAE, PointTransformer) & 45.26 \\
    Mask Point-MAE (Point-MAE, MaskTransformer) & 43.56 \\
    ShapeNet55-34 (PCP-MAE, MaskTransformer) & 46.58 \\
    Random (PointTransformer) & 44.45 \\
    \bottomrule
  \end{tabular}%
  }
\end{table}

\textbf{Finding 1: Unfreezing dramatically improves captioning.} Gains range from +13.39 (V1 hybrid) to +18.38 (V2) and +17.39 (ShapeNet55-34). V1 hybrid unfrozen closes 72\% of the gap to Baseline (36.24 $\rightarrow$ 49.63 vs.\ Baseline 53.37).

\textbf{Finding 2: Random initialization is remarkably competitive.} Random unfrozen (44.45) surpasses V2 unfrozen (43.94) and trails V1 hybrid unfrozen (49.63) by only 5.18 points. This confirms that generative captioning tasks are more resilient to encoder quality than discriminative classification.

\textbf{Finding 3: The center prediction bias reverses the advantage compared to classification.} In classification, PCP-MAE slightly helps; in captioning, Point-MAE (pure reconstruction) consistently outperforms PCP-MAE (with center prediction) across both architectures (frozen: 33.46 vs.\ 25.56 for PointTransformer; 28.62 vs.\ 36.24 for MaskTransformer---note the opposite trend). The center prediction bias toward global structure appears to interfere with the fine-grained geometric detail needed for descriptive captioning.

\subsection{Ablation Studies}
\label{sec:ablations}

\subsubsection{Unfreezing is Necessary but Not Sufficient}

Unfreezing consistently improves all metrics, but the improvement magnitude varies by task and architecture:

\begin{table}[t]
  \centering
  \small
  \caption{Summary of frozen $\rightarrow$ unfrozen gains across metrics.}
  \label{tab:unfreeze_summary}
  \resizebox{\columnwidth}{!}{%
  \begin{tabular}{lcc}
    \toprule
    \textbf{Metric} & \textbf{Avg frozen $\rightarrow$ unfrozen gain} & \textbf{Random unfrozen vs.\ Baseline gap} \\
    \midrule
    Open-vocab avg & +10.9pp & $-$14.00pp \\
    ModelNet40 clean avg & +1.9pp & $-$50.24pp \\
    Captioning & +15.2 & $-$8.92 \\
    \bottomrule
  \end{tabular}%
  }
\end{table}

The larger captioning gains (+15.2 on average) suggest generative tasks are where unfreezing most effectively bridges the semantic gap. The small ModelNet40 gain (+1.9pp on average) confirms that closed-set classification is fundamentally limited by the encoder's representational capacity.

\textbf{Crucially, unfreezing alone cannot close the gap for PointTransformer-based self-supervised encoders:} V2 unfrozen (52.50\%) is identical to Random unfrozen (52.50\%), indicating no benefit from pre-training.

\subsubsection{Architecture $\times$ Pre-training Objective Crossover Interaction}

\begin{figure*}[t]
  \centering
  \includegraphics[width=\textwidth]{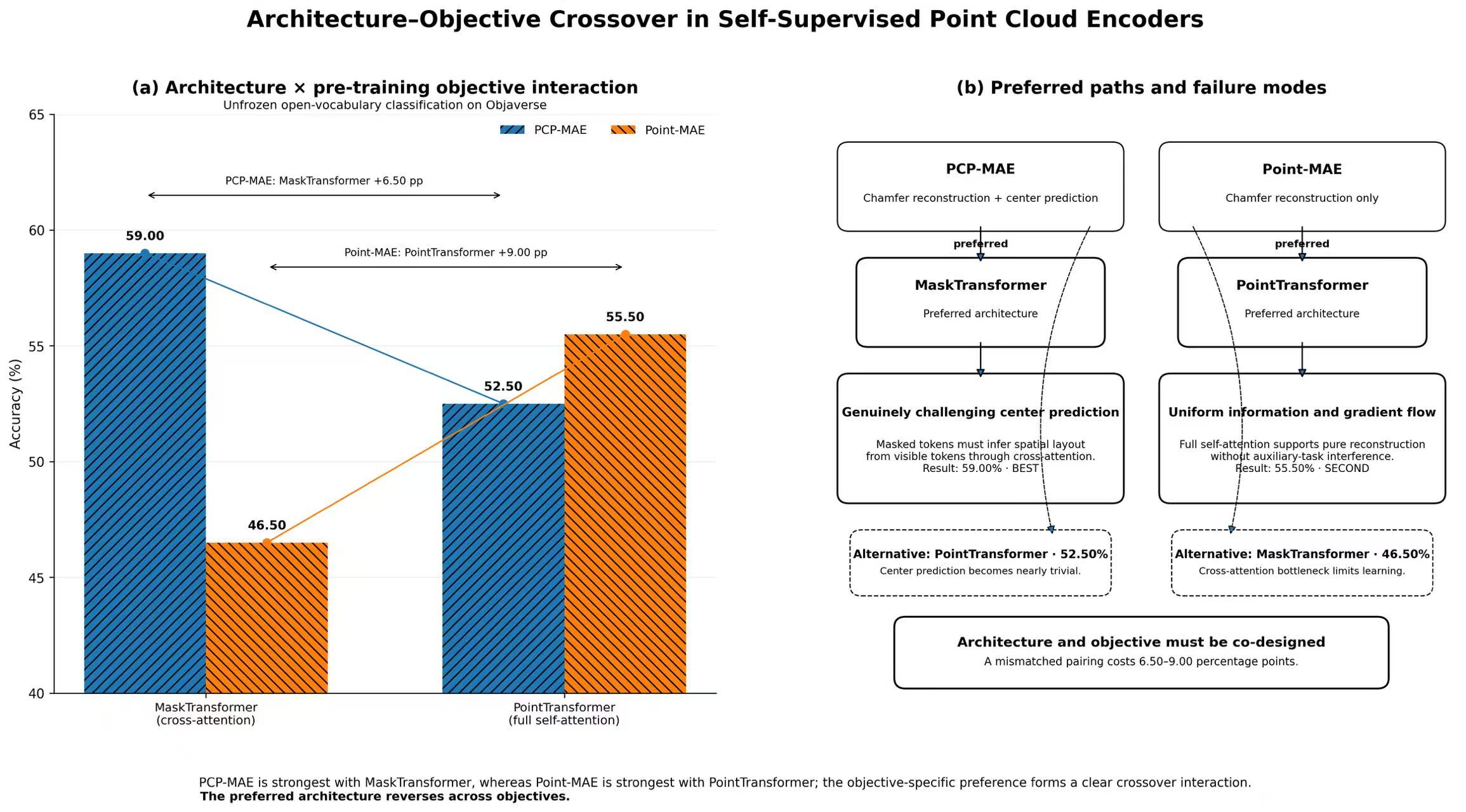}
  \caption{\textbf{Architecture $\times$ Pre-training Objective Crossover Interaction.} A 2$\times$2 matrix with 4 condition bars: PCP-MAE+MaskTransformer=59.00\%, PCP-MAE+PointTransformer=52.50\%, Point-MAE+PointTransformer=55.50\%, Point-MAE+MaskTransformer=46.50\%. The CrossoverPattern node documents the inversion: optimal architecture flips depending on pre-training objective. Four CrossoverArrows provide detailed mechanism explanations (green for wins, red for losses). The MechanismDetail node explains why center prediction has opposite effects based on how masked token features are computed (cross-attention in MaskTransformer vs.\ self-attention in PointTransformer). Cross-task validation confirms the crossover is consistent across all 3 tasks. Includes 4 practical recommendations for 3D-LLM design.}
  \label{fig:crossover}
\end{figure*}

This is the most novel finding of our study. Table~\ref{tab:crossover} presents the interaction matrix for unfrozen open-vocabulary classification:

\begin{table}[t]
  \centering
  \small
  \caption{Open-vocabulary classification accuracy (\%) showing the crossover interaction between architecture and pre-training objective (unfrozen).}
  \label{tab:crossover}
  \resizebox{\columnwidth}{!}{%
  \begin{tabular}{lcc}
    \toprule
    \textbf{Architecture} & \textbf{PCP-MAE (with center pred)} & \textbf{Point-MAE (pure recon)} \\
    \midrule
    MaskTransformer & \textbf{59.00\%} (V1 hybrid) & 46.50\% (Mask Point-MAE) \\
    PointTransformer & 52.50\% (V2) & \textbf{55.50\%} (Point-MAE) \\
    \bottomrule
  \end{tabular}%
  }
\end{table}

The pattern is striking:
\begin{itemize}
\item \textbf{PCP-MAE benefits from MaskTransformer (+6.50pp)}
\item \textbf{Point-MAE benefits from PointTransformer (+9.00pp)}
\end{itemize}

This crossover can be understood by examining precisely where the center prediction head attaches. In both architectures, the center prediction head ($\texttt{pred\_pos\_proj}$) is applied directly to the \textbf{masked token features} $x_{\text{mask}}$ produced by the encoder. The critical difference lies in how those masked token features are computed:

\begin{itemize}
\item \textbf{MaskTransformer} maintains two separate token groups throughout its transformer blocks: $x_{\text{vis}}$ (visible) undergoes self-attention among visible patches, while $x_{\text{mask}}$ (masked) receives \textbf{cross-attention from masked to visible patches only}---it never sees other masked tokens. To predict its own center, $x_{\text{mask}}$ must \textit{infer} it indirectly by querying visible patches and reasoning about their spatial arrangement. This genuinely challenging task forces the encoder to learn rich relational representations of visible geometry. The auxiliary center loss thereby reinforces a representation that matches MaskTransformer's cross-attention strength: reasoning about spatial relationships between a query token and a context set.

\item \textbf{PointTransformer} (the encoder used in V2, Point-MAE, and Random) processes all $G$ patch tokens plus a [cls] token through \textbf{full self-attention} over the entire sequence, then \textit{post-hoc} splits into $x_{\text{vis}}$ and $x_{\text{mask}}$ by indexing. Consequently, $x_{\text{mask}}$ has already attended to every visible token (and every other masked token) via standard self-attention, so it possesses rich global context before the center prediction head is reached. Predicting the center becomes nearly trivial---the network can simply look up positional information already present in the self-attended features. The added center loss thus provides little training signal and, in the worst case, biases the backbone toward learning easily regressed global positions rather than locally discriminative geometry. When coupled with PointTransformer, removing the center loss (i.e., Point-MAE) yields strictly better performance (55.50\% vs.\ 52.50\%).
\end{itemize}

Conversely, when the simpler reconstruction objective is paired with MaskTransformer, the cross-attention bottleneck between $x_{\text{vis}}$ and $x_{\text{mask}}$ may restrict gradient flow to the visible-path encoder, explaining why Mask Point-MAE underperforms Point-MAE on PointTransformer (46.50\% vs.\ 55.50\%).

This crossover interaction has significant practical implications: \textbf{choosing the wrong architecture for a given pre-training objective can cost 6.50--9.00pp in accuracy.}

\subsubsection{Pre-training vs.\ Random Initialization}

\begin{table*}[t]
  \centering
  \small
  \caption{Performance comparison of pre-trained vs.\ random initialization (all unfrozen).}
  \label{tab:pretrain_vs_random}
  \begin{tabular}{lccc}
    \toprule
    \textbf{Variant (all unfrozen)} & \textbf{Open-vocab avg} & \textbf{Captioning} & \textbf{ModelNet40 clean avg} \\
    \midrule
    Random & 52.50\% & 44.45 & 12.21\% \\
    V2 (PCP-MAE, PointTransformer) & 52.50\% & 43.94 & 12.36\% \\
    Point-MAE (PointTransformer) & \textbf{55.50\%} & \textbf{45.26} & \textbf{16.56\%} \\
    ShapeNet55-34 (MaskTransformer) & 53.50\% & 46.58 & 16.30\% \\
    V1 hybrid (MaskTransformer) & \textbf{59.00\%} & \textbf{49.63} & \textbf{17.83\%} \\
    \bottomrule
  \end{tabular}
\end{table*}

\textbf{Finding: The marginal benefit of self-supervised pre-training over random initialization varies by architecture and objective.} For PointTransformer, Point-MAE (pure reconstruction) shows a consistent advantage over random initialization: +3.00pp on open-vocab (55.50\% vs.\ 52.50\%), +4.35pp on ModelNet40 clean (16.56\% vs.\ 12.21\%), and +0.81 on captioning. However, PCP-MAE on PointTransformer (V2) shows no gain, performing identically to Random (52.50\%). For MaskTransformer, PCP-MAE (V1 hybrid) provides a notable advantage (+6.50pp over Random), while Point-MAE (Mask Point-MAE) underperforms Random. Only the MaskTransformer + PCP-MAE combination (V1 hybrid) provides a consistent and substantial advantage across all metrics.

This finding suggests that \textbf{the dominant source of performance in MiniGPT-3D is the cascaded four-stage training pipeline and the strong 2D-LLM priors, but Point-MAE pre-training on PointTransformer provides a meaningful---if modest---additional benefit (3--4pp) over random initialization.}

\subsubsection{Pre-training Dataset: Scale vs.\ Quality}

ShapeNet55-34 ($\sim$50K CAD models, 3-channel xyz only) performs competitively with Objaverse (660K real-world objects, 6-channel xyz+rgb):

\begin{table}[t]
  \centering
  \small
  \caption{ShapeNet55-34 vs.\ best Objaverse variant comparison.}
  \label{tab:dataset_comparison}
  \resizebox{\columnwidth}{!}{%
  \begin{tabular}{lcc}
    \toprule
    \textbf{Setting} & \textbf{ShapeNet55-34} & \textbf{Best Objaverse variant} \\
    \midrule
    Frozen open-vocab & 50.00\% & 49.25\% (V1 hybrid) \\
    Unfrozen open-vocab & 53.50\% & 59.00\% (V1 hybrid) \\
    Unfrozen captioning & \textbf{46.58} & 49.63 (V1 hybrid) \\
    Unfrozen ModelNet40 clean & \textbf{16.30\%} & 17.83\% (V1 hybrid) \\
    \bottomrule
  \end{tabular}%
  }
\end{table}

Despite 13$\times$ fewer samples and no color, ShapeNet55-34 unfrozen comes within 5.50pp of V1 hybrid on classification and within 3.05 on captioning, while surpassing or matching V2 and Random. This indicates that \textbf{high-quality CAD models from ShapeNet provide geometric signals comparable in utility to large-scale real-world scans}, especially when the downstream pipeline can adapt the encoder.

\subsubsection{Center Prediction: Mixed Effects}

\begin{table}[t]
  \centering
  \small
  \caption{Center prediction effects by architecture (unfrozen).}
  \label{tab:center_pred}
  \resizebox{\columnwidth}{!}{%
  \begin{tabular}{lcccc}
    \toprule
    \textbf{Task} & \multicolumn{2}{c}{\textbf{PointTransformer}} & \multicolumn{2}{c}{\textbf{MaskTransformer}} \\
    & \textbf{PCP-MAE} & \textbf{Point-MAE} & \textbf{PCP-MAE} & \textbf{Point-MAE} \\
    \midrule
    Open-vocab avg & 52.50\% & \textbf{55.50\%} & \textbf{59.00\%} & 46.50\% \\
    Captioning & 43.94 & \textbf{45.26} & \textbf{49.63} & 43.56 \\
    ModelNet40 clean avg & 12.36\% & \textbf{16.56\%} & \textbf{17.83\%} & 12.87\% \\
    \bottomrule
  \end{tabular}%
  }
\end{table}

With PointTransformer, \textbf{Point-MAE (no center prediction) outperforms PCP-MAE on all three tasks.} With MaskTransformer, \textbf{PCP-MAE (with center prediction) outperforms Point-MAE on all three tasks.} This asymmetry aligns with the crossover interaction described in Section~4.2.2: center prediction is only beneficial when paired with an architecture designed for cross-patch interaction.

\subsection{Encoder Feature Space Analysis}
\label{sec:feature_analysis}

\subsubsection{Cross-Encoder Feature Similarity}

\begin{table*}[t]
  \centering
  \small
  \caption{Cosine similarity between variant and Baseline encoder features (measured on 2,468 ModelNet40 samples). Note: Point-MAE data uses the updated max\_samples=2468 setting for consistency with other variants.}
  \label{tab:cross_encoder}
  \begin{tabular}{lccc}
    \toprule
    \textbf{Encoder} & \textbf{cls} & \textbf{global} & \textbf{router} \\
    \midrule
    Pre-V1 (PCP-MAE, MaskTransformer, no cls) & 0.0541 & 0.3917 & 0.2711 \\
    V1 hybrid (PCP-MAE, MaskTransformer, cls copied) & 0.0346 & 0.3915 & 0.2617 \\
    V2 (PCP-MAE, PointTransformer, native cls) & 0.0289 & 0.4722 & 0.2998 \\
    Point-MAE (Point-MAE, PointTransformer, native cls) & $-$0.0125 & 0.4330 & 0.2606 \\
    Mask Point-MAE (Point-MAE, MaskTransformer, no cls) & $-$0.0019 & 0.1492 & 0.0779 \\
    \bottomrule
  \end{tabular}
\end{table*}

\textbf{Finding 1: cls features are nearly orthogonal to the Baseline for all variants (cosine $\leq$ 0.054).} Even V1 hybrid, which copies the Baseline's cls\_token and cls\_pos, achieves only 0.0346 cosine---confirming that cls output is determined by the full backbone's self-attention dynamics. This explains why cls parameter copying is ineffective.

\textbf{Finding 2: Global features show moderate alignment (0.15--0.47).} V2's PointTransformer achieves the highest global cosine (0.4722), indicating that architecture matching with the Baseline helps global feature alignment. However, V2's downstream performance is worse than V1 hybrid's, proving that \textbf{global feature alignment is not a reliable predictor of downstream 3D-LLM performance.}

\textbf{Finding 3: Mask Point-MAE shows the lowest alignment with the Baseline} (global 0.1492, router 0.0779), correlating with its poor downstream performance.

\textbf{Finding 4: Point-MAE's global feature similarity (0.4330) is comparable to V2 (0.4722)} despite its higher downstream performance, further confirming that feature alignment alone does not determine 3D-LLM task performance.

\subsubsection{kNN Classification on ModelNet40}

\begin{table*}[t]
  \centering
  \small
  \caption{kNN accuracy on ModelNet40 using frozen features. The baseline row shows the multi-modal encoder's own kNN performance for reference; variant rows show each self-supervised encoder's actual kNN performance.}
  \label{tab:knn}
  \begin{tabular}{lccc}
    \toprule
    \textbf{Encoder} & \textbf{cls} & \textbf{global} & \textbf{router} \\
    \midrule
    Baseline (ULIP-2 Point-BERT) & 72.67\% & 79.35\% & 78.54\% \\
    Pre-V1 (PCP-MAE, MaskTransformer) & 60.93\% & 76.32\% & 72.27\% \\
    V1 hybrid (PCP-MAE, MaskTransformer) & 60.93\% & 77.33\% & 67.41\% \\
    V2 (PCP-MAE, PointTransformer) & 63.56\% & 74.49\% & 70.85\% \\
    Point-MAE (Point-MAE, PointTransformer) & 61.74\% & 68.22\% & 68.22\% \\
    Mask Point-MAE (Point-MAE, MaskTransformer) & 60.73\% & 61.94\% & 61.13\% \\
    \bottomrule
  \end{tabular}
\end{table*}

All self-supervised encoders achieve 60--64\% kNN accuracy, confirming that \textbf{self-supervised features encode discriminative geometric information in their own representational space}---the issue is misalignment with the space that the downstream LLM modules expect, not absence of discriminability. However, the gap between Baseline (72--80\%) and the best self-supervised variant (61--77\%) is substantial (5--13pp).

The gap between cls kNN and global kNN varies by variant: PointTransformer-based variants (V2, Point-MAE) show $\sim$10pp gap favoring global features, while MaskTransformer-based variants (Pre-V1, V1 hybrid) show a smaller gap ($\sim$3pp). This suggests that the cls token is not the optimal representation for PointTransformer-based encoders, supporting the use of global pooling as an alternative.

\subsubsection{Intra-Class vs.\ Inter-Class Separability}

\begin{table}[t]
  \centering
  \small
  \caption{Intra-class vs.\ inter-class cosine similarity and class separability margin.}
  \label{tab:separability}
  \resizebox{\columnwidth}{!}{%
  \begin{tabular}{lccc}
    \toprule
    \textbf{Encoder} & \textbf{Intra-class} & \textbf{Inter-class} & \textbf{Margin} \\
    \midrule
    Baseline (cls) & 0.8465 & 0.7303 & 0.1162 \\
    V1 hybrid (cls) & 0.8349 & 0.7457 & 0.0892 \\
    V2 (cls) & 0.9162 & 0.8293 & 0.0869 \\
    \bottomrule
  \end{tabular}%
  }
\end{table}

\textbf{Finding: Self-supervised encoders have $\sim$23--26\% lower class separability margin (0.087--0.089 vs.\ 0.116).} The V2 encoder has \textit{higher} intra-class similarity than the Baseline (0.9162 vs.\ 0.8465) but also much higher inter-class similarity (0.8293 vs.\ 0.7303), collapsing the margin. This confirms that geometric self-supervision learns to map all objects into a compact feature space where within-class and between-class variations are poorly distinguished---a direct consequence of the reconstruction objective that treats all points equally regardless of semantic category.

The $\sim$26\% margin reduction explains the $\sim$70\% absolute drop in ModelNet40 closed-set accuracy (62\% $\rightarrow$ 18\%). Small differences in initial feature separability are amplified through the frozen alignment layers, and even unfreezing cannot fully recover discriminability.

\section{Discussion}
\label{sec:discussion}

\subsection{When Can Self-Supervised Encoders Replace Multi-Modal Pre-training?}

Our results paint a nuanced picture across three dimensions:

\textbf{For open-vocabulary tasks (classification and captioning):} Self-supervised encoders can approach multi-modal encoder performance, especially with MaskTransformer + PCP-MAE pre-training and encoder unfreezing (V1 hybrid achieves 88\% of Baseline on classification, 93\% on captioning). \textbf{However, this success is largely attributable to the downstream pipeline rather than the pre-training itself}, as a randomly initialized encoder also reaches 78\% and 83\% of baseline on these tasks.

\textbf{For closed-set classification:} Self-supervised encoders fail categorically---all variants plateau at 13--18\% vs.\ 62\% baseline. This ceiling is independent of architecture, objective, data scale, and fine-tuning. Multi-modal pre-training's ability to embed semantic category boundaries is the decisive factor.

\textbf{On the marginal value of self-supervised pre-training:} For PointTransformer architecture, Point-MAE pre-training provides a small but consistent benefit over random initialization (+3.00pp open-vocab, +4.35pp ModelNet40), while PCP-MAE shows no benefit. For MaskTransformer, the pattern inverts: PCP-MAE provides a strong advantage (+6.50pp), while Point-MAE is detrimental. Only the correct architecture-objective pairing yields meaningful gains.

\subsection{Cost-Efficiency Analysis}

A key motivation for exploring self-supervised encoders is reducing the computational and data costs of multi-modal pre-training. From a practical cost perspective:

\begin{itemize}
\item \textbf{ULIP-2 multi-modal pre-training} requires 8$\times$ NVIDIA A100 GPUs for 3--4 days, with total storage requirements of $\sim$1.2TB (Objaverse point clouds, rendered images, and text descriptions). The data collection and curation pipeline for tri-modal alignment is itself a significant engineering effort.
\item \textbf{Self-supervised pre-training (PCP-MAE / Point-MAE)} requires only a single RTX 3090 GPU for approximately 2 days (Objaverse 660K, 24--35 epochs), with storage requirements of $\sim$800GB (point clouds only, no images or text). Data acquisition is reduced to downloading public point cloud datasets without additional modality processing.
\end{itemize}

This represents an $\sim$85\% reduction in GPU compute cost and $\sim$33\% reduction in storage, while completely eliminating the need for image rendering and text description pipelines. Even the best self-supervised variant (V1 hybrid unfrozen) achieves up to 88\% of Baseline classification and 93\% of Baseline captioning performance---a favorable cost-performance trade-off for many practical applications.

\subsection{Applicability Boundary}

Our finding that ``random initialization approaches pre-trained performance'' should not be over-generalized. This conclusion specifically applies to \textbf{efficient 3D-LLM architectures that leverage strong 2D-LLM priors and cascaded modality alignment training} (MiniGPT-3D's four-stage pipeline). In such architectures, the LLM already possesses rich visual-semantic knowledge from its 2D pre-training, and the cascaded alignment layers can effectively guide the encoder to learn compatible representations during fine-tuning. For 3D-LLMs that lack 2D priors and align point clouds directly with text from scratch (\eg, training the entire pipeline from random initialization), the benefits of self-supervised encoder pre-training are likely to be substantially larger.

\subsection{Practical Recommendations}

\begin{enumerate}
\item \textbf{Do not assume self-supervised pre-training will help if you plan to fine-tune.} Our results show that with PointTransformer (the default in most 3D-LLMs), PCP-MAE pre-training provides no benefit over random initialization when the encoder is trainable, though Point-MAE provides a small benefit ($\sim$3pp). Computational resources may be better allocated to the downstream training pipeline.

\item \textbf{If using PCP-MAE pre-training, pair it with MaskTransformer.} The center prediction and cross-attention mechanisms produce synergistic effects (59.00\% vs.\ 52.50\%).

\item \textbf{If using pure masked reconstruction (Point-MAE), pair it with PointTransformer.} The simpler self-attention architecture better serves the reconstruction objective (55.50\% vs.\ 46.50\%).

\item \textbf{Data quality can substitute for data scale.} ShapeNet55-34's 50K high-quality CAD models match or exceed Objaverse's 660K samples when pre-training a MaskTransformer.

\item \textbf{For closed-set classification tasks, multi-modal pre-training remains irreplaceable.} No combination of geometric self-supervision and downstream fine-tuning approached the Baseline's performance on ModelNet40.
\end{enumerate}

\subsection{Limitations and Future Work}

\textbf{Limitations.} (1) All self-supervised encoders were pre-trained for fewer than the full 300 scheduled epochs (24-35 epochs on Objaverse; 71 epochs on ShapeNet55-34). Early stopping was applied when the reconstruction loss entered a plateau, but it is possible that longer pre-training could yield different results. (2) The study is limited to object-level point clouds on a single 3D-LLM pipeline (MiniGPT-3D). (3) Evaluations use the Qwen-Flash API, which differs from GPT-4 evaluations in prior work. (4) The MiniGPT-3D pipeline uses a 2.7B Phi-2 LLM; larger LLMs with stronger semantic reasoning may narrow the self-supervised gap. (5) ShapeNet55-34 has higher category overlap with ModelNet40 and cleaner geometry than Objaverse, which may partially explain its competitive performance despite smaller scale---the effects of dataset quality and dataset scale have not been fully decoupled.

\textbf{Future work.} (1) Pre-training self-supervised encoders to full convergence (300 epochs) to measure the upper-bound benefit. (2) Investigating whether pre-training benefits re-emerge with larger LLM backbones ($>$7B parameters). (3) Exploring hybrid objectives that combine geometric reconstruction with lightweight semantic signals (\eg, weak supervision from category labels or text descriptions). (4) Extending the study to scene-level 3D-LLMs and other efficient 3D-LLM architectures.

\section{Conclusion}
\label{sec:conclusion}

We presented a systematic empirical study on replacing multi-modal pre-trained point cloud encoders with self-supervised alternatives in efficient 3D-LLM pipelines. Through 7 encoder initialization / pre-training configurations evaluated under 2 fine-tuning strategies (12 experimental groups total), 2 architectures, 3 pre-training objectives, and 2 pre-training datasets, we characterized the conditions under which self-supervised encoders can serve as cost-effective alternatives.

Our central findings are: \textbf{(1) The MiniGPT-3D pipeline is remarkably robust to encoder quality}---a randomly initialized encoder achieves competitive performance when fine-tuned, matching some pre-trained variants. \textbf{(2) Architecture and pre-training objective exhibit a strong crossover interaction} that must be carefully considered: PCP-MAE with MaskTransformer (59.00\%) far outperforms all other combinations, while Point-MAE with MaskTransformer (46.50\%) underperforms Point-MAE with PointTransformer (55.50\%). \textbf{(3) Closed-set classification is a fundamental limitation of all current geometric self-supervision methods}, plateauing at $\sim$18\% regardless of configuration, due to the absence of semantic category boundaries in the learned feature space.

We hope these findings guide practitioners in building cost-effective 3D-LLMs and inform the design of next-generation self-supervised pre-training methods that can bridge the remaining gap with multi-modal representation learning.

{
    \small
    \bibliographystyle{ieeenat_fullname}
    \bibliography{main}

@String(CVPR= {IEEE Conf. Comput. Vis. Pattern Recog.})

@String(ECCV= {Eur. Conf. Comput. Vis.})

@String(ICLR = {Int. Conf. Learn. Represent.})

@String(CVPR  = {CVPR})

@String(ECCV  = {ECCV})

@String(ICLR  = {ICLR})

@article{touvron2023llama,
  title         = {{LLaMA}: Open and Efficient Foundation Language Models},
  author        = {Hugo Touvron and Thibaut Lavril and Gautier Izacard and Xavier Martinet and Marie-Anne Lachaux and Timothée Lacroix and Baptiste Rozière and Naman Goyal and Eric Hambro and Faisal Azhar and Aurelien Rodriguez and Armand Joulin and Edouard Grave and Guillaume Lample},
  journal       = {arXiv preprint},
  eprint        = {2302.13971},
  archivePrefix = {arXiv},
  primaryClass  = {cs.CL},
  year          = {2023},
  doi           = {10.48550/arXiv.2302.13971}
}

@article{openai2023gpt4,
  title         = {{GPT}-4 Technical Report},
  author        = {OpenAI},
  journal       = {arXiv preprint},
  eprint        = {2303.08774},
  archivePrefix = {arXiv},
  primaryClass  = {cs.CL},
  year          = {2023},
  doi           = {10.48550/arXiv.2303.08774}
}

@inproceedings{li2023blip2,
  title     = {{BLIP}-2: Bootstrapping Language-Image Pre-training with Frozen Image Encoders and Large Language Models},
  author    = {Junnan Li and Dongxu Li and Silvio Savarese and Steven Hoi},
  booktitle = {Proceedings of the 40th International Conference on Machine Learning (ICML)},
  volume    = {202},
  pages     = {20351--20383},
  publisher = {PMLR},
  year      = {2023}
}

@inproceedings{dai2023instructblip,
  title     = {{InstructBLIP}: Towards General-purpose Vision-Language Models with Instruction Tuning},
  author    = {Wenliang Dai and Junnan Li and Dongxu Li and Anthony Tiong and Junqi Zhao and Weisheng Wang and Boyang Li and Pascale Fung and Steven Hoi},
  booktitle = {Advances in Neural Information Processing Systems (NeurIPS)},
  volume    = {36},
  year      = {2023}
}

@article{qi2024shapellm,
  title         = {{ShapeLLM}: Universal 3D Object Understanding for Embodied Interaction},
  author        = {Zekun Qi and Runpei Dong and Shaochen Zhang and Haoran Geng and Chunrui Han and Zheng Ge and Li Yi and Kaisheng Ma},
  journal       = {arXiv preprint},
  eprint        = {2402.17766},
  archivePrefix = {arXiv},
  primaryClass  = {cs.CV},
  year          = {2024},
  doi           = {10.48550/arXiv.2402.17766}
}

@article{xu2023pointllm,
  title         = {{PointLLM}: Empowering Large Language Models to Understand Point Clouds},
  author        = {Runsen Xu and Xiaolong Wang and Tai Wang and Yilun Chen and Jiangmiao Pang and Dahua Lin},
  journal       = {arXiv preprint},
  eprint        = {2308.16911},
  archivePrefix = {arXiv},
  primaryClass  = {cs.CV},
  year          = {2023},
  doi           = {10.48550/arXiv.2308.16911}
}

@article{tang2024minigpt3d,
  title         = {{MiniGPT}-3{D}: Efficiently Aligning 3D Point Clouds with Large Language Models using 2D Priors},
  author        = {Yuan Tang and Xu Han and Xianzhi Li and Qiao Yu and Yixue Hao and Long Hu and Min Chen},
  journal       = {arXiv preprint},
  eprint        = {2405.01413},
  archivePrefix = {arXiv},
  primaryClass  = {cs.CV},
  year          = {2024},
  doi           = {10.48550/arXiv.2405.01413},
  note          = {Also appeared in ACM MM 2024}
}

@inproceedings{xue2024ulip2,
  title     = {{ULIP}-2: Towards Scalable Multimodal Pre-training for 3D Understanding},
  author    = {Le Xue and Ning Yu and Shu Zhang and Artemis Panagopoulou and Junnan Li and Roberto Martín-Martín and Jiajun Wu and Caiming Xiong and Ran Xu and Juan Carlos Niebles and Silvio Savarese},
  booktitle = {Proceedings of the IEEE/CVF Conference on Computer Vision and Pattern Recognition (CVPR)},
  pages     = {22945--22955},
  year      = {2024}
}

@inproceedings{yu2022pointbert,
  title     = {{Point-BERT}: Pre-training 3D Point Cloud Transformers with Masked Point Modeling},
  author    = {Xumin Yu and Lulu Tang and Yongming Rao and Tiejun Huang and Jie Zhou and Jiwen Lu},
  booktitle = {Proceedings of the IEEE/CVF Conference on Computer Vision and Pattern Recognition (CVPR)},
  pages     = {19313--19322},
  year      = {2022},
  doi       = {10.1109/CVPR52729.2022.01865}
}

@inproceedings{pang2022pointmae,
  title     = {Masked Autoencoders for Point Cloud Self-supervised Learning},
  author    = {Yatian Pang and Wenxiao Wang and Francis E. H. Tay and Wei Liu and Yonghong Tian and Li Yuan},
  booktitle = {Proceedings of the 17th European Conference on Computer Vision (ECCV)},
  pages     = {591--608},
  publisher = {Springer},
  year      = {2022},
  doi       = {10.1007/978-3-031-19806-9_34}
}

@article{zhang2024pcpmae,
  title         = {{PCP-MAE}: Learning to Predict Centers for Point Masked Autoencoders},
  author        = {Xiangdong Zhang and Shaofeng Zhang and Junchi Yan},
  journal       = {arXiv preprint},
  eprint        = {2408.08753},
  archivePrefix = {arXiv},
  primaryClass  = {cs.CV},
  year          = {2024},
  doi           = {10.48550/arXiv.2408.08753},
  note          = {Also appeared in NeurIPS 2024}
}

@inproceedings{hong2023three,
  title     = {{3D-LLM}: Injecting the 3D World into Large Language Models},
  author    = {Yining Hong and Haoyu Zhen and Pehao Chen and Shuhong Zheng and Yilun Du and Zhenfang Chen and Chuang Gan},
  booktitle = {Advances in Neural Information Processing Systems (NeurIPS)},
  volume    = {36},
  year      = {2023}
}

@article{chen2023ll3da,
  title         = {{LL3DA}: Visual Interactive Instruction Tuning for Omni-3D Understanding, Reasoning, and Planning},
  author        = {Sijin Chen and Xin Chen and Chi Zhang and Mingsheng Li and Gang Yu and Hao Fei and Hongyuan Zhu and Jiayuan Fan and Tao Chen},
  journal       = {arXiv preprint},
  eprint        = {2311.18651},
  archivePrefix = {arXiv},
  primaryClass  = {cs.CV},
  year          = {2023},
  doi           = {10.48550/arXiv.2311.18651}
}

@inproceedings{hu2022lora,
  title     = {{LoRA}: Low-Rank Adaptation of Large Language Models},
  author    = {Edward J. Hu and Yelong Shen and Phillip Wallis and Zeyuan Allen-Zhu and Yuanzhi Li and Shean Wang and Lu Wang and Weizhu Chen},
  booktitle = {International Conference on Learning Representations (ICLR)},
  year      = {2022}
}

@article{yuan2023tinygptv,
  title         = {{TinyGPT-V}: Efficient Multimodal Large Language Model via Small Backbones},
  author        = {Zhengqing Yuan and Zhaoxu Li and Lichao Sun},
  journal       = {arXiv preprint},
  eprint        = {2312.16862},
  archivePrefix = {arXiv},
  primaryClass  = {cs.CV},
  year          = {2023},
  doi           = {10.48550/arXiv.2312.16862},
  note          = {Also appeared in ICML 2024 Workshop}
}

@article{chu2023mobilevlm,
  title         = {{MobileVLM}: A Fast, Reproducible and Strong Vision Language Assistant for Mobile Devices},
  author        = {Xiangxiang Chu and Limeng Qiao and Xinyang Lin and Shuang Xu and Yang Yang and Yiming Hu and Fei Wei and Xinyu Zhang and Bo Zhang and Xiaolin Wei and Chunhua Shen},
  journal       = {arXiv preprint},
  eprint        = {2312.16886},
  archivePrefix = {arXiv},
  primaryClass  = {cs.CV},
  year          = {2023},
  doi           = {10.48550/arXiv.2312.16886}
}
}

\clearpage
\maketitlesupplementary

\section{Implementation and Reproduction Details}
\label{sec:appendix}

\subsection{Baseline Reproduction Results}
\label{sec:baseline_repro}

We fully reproduced the MiniGPT-3D Baseline (ULIP-2 pre-trained Point-BERT encoder) by re-running the complete four-stage training pipeline from scratch. Table~\ref{tab:baseline_repro} compares our reproduced results with the official weights evaluation.

\begin{table}[ht]
  \centering
  \small
  \caption{Comparison of official weights evaluation vs.\ our reproduction results.}
  \label{tab:baseline_repro}
  \resizebox{\columnwidth}{!}{%
  \begin{tabular}{lcc}
    \toprule
    \textbf{Task} & \textbf{Official Weights} & \textbf{Our Reproduction} \\
    \midrule
    Open-vocab classification (Prompt 0) & 66.00\% & 67.00\% \\
    Open-vocab classification (Prompt 1) & 68.00\% & 66.00\% \\
    ModelNet40 clean (Prompt 0) & 62.66\% & 63.83\% \\
    ModelNet40 clean (Prompt 1) & 60.75\% & 61.06\% \\
    Captioning score & 53.57 & 53.37 \\
    \bottomrule
  \end{tabular}%
  }
\end{table}

The results are closely consistent, confirming that our reproduction is faithful to the original. All variant experiments in the main paper are compared against this reproduced Baseline.

\subsection{Gradient Blocking in PCP-MAE Pre-training}
\label{sec:gradient_blocking}

During initial PCP-MAE pre-training on Objaverse, we encountered a critical gradient blocking issue. When enabling automatic mixed precision (AMP), the Chamfer reconstruction loss stagnated at approximately 1,000 for over 30 epochs. Investigation revealed that during AMP debugging, the following code pattern had been inadvertently introduced:

{\footnotesize
\begin{verbatim}
with torch.no_grad():
    rebuild_points_fp32 = rebuild_points.float()
    gt_coords_fp32 = gt_coords.float()
    loss1 = self.loss_func(rebuild_points_fp32,
                           gt_coords_fp32)
\end{verbatim}}

The \texttt{torch.no\_grad()} context completely detached the Chamfer reconstruction loss from the computation graph, preventing gradients from flowing back to the encoder and decoder. Removing this block restored convergence within a single epoch, with the Chamfer loss dropping from $\sim$1,000 to below 1.0.

\subsection{AMP Numerical Stability}
\label{sec:amp_stability}

Training with float16 AMP on the RTX 3090 caused NaN/Inf values during the late stages of training (around epoch 15--38 depending on the variant). Analysis identified two root causes:

\begin{enumerate}
\item \textbf{Transformer operations in float16 precision}: Large matrix multiplications, Softmax, and LayerNorm operations in Transformer blocks are prone to overflow in float16.
\item \textbf{Chamfer distance computation}: The Chamfer distance involves squared Euclidean distances that can produce large values exceeding the float16 dynamic range.
\end{enumerate}

The following measures were applied to resolve the issue:

\begin{itemize}
\item \textbf{Switched from float16 to bfloat16 AMP}: The RTX 3090 supports bfloat16, which provides the same dynamic range as float32 while retaining the memory and throughput benefits of reduced precision.
\item \textbf{Forced Chamfer distance computation in float32}: The Chamfer distance loss function was explicitly cast to float32 precision regardless of the AMP setting.
\item \textbf{Gradient clipping}: Applied gradient clipping with a maximum norm of 1.0 to prevent gradient explosion.
\end{itemize}

These modifications ensured stable training without NaN/Inf occurrences for all remaining experiments.

\subsection{API Model Deprecation and Migration}
\label{sec:api_migration}

The original MiniGPT-3D project relied on the Qwen2-72B-Instruct API for evaluation. During our study, this API was decommissioned, requiring migration to the Alibaba Cloud Bailian Qwen-Flash API. Key differences and adaptations:

\begin{itemize}
\item The Qwen2-72B-Instruct model is no longer available on the Bailian platform; the most recent available model was Qwen2.5-VL-32B-Instruct (which itself was scheduled for deprecation around May 13, 2026).
\item We migrated all evaluation scripts to use the \texttt{qwen-flash} model endpoint.
\item API parameters (temperature, max tokens, prompt templates) were kept consistent across all evaluation runs.
\item Pricing parameters in the evaluation scripts were updated to match Qwen-Flash billing rates.
\item The evaluation logic was modified to dynamically adapt to different model endpoints, enabling future model changes without code modification.
\end{itemize}

\subsection{Pre-training Hyperparameters}
\label{sec:hyperparams}

\begin{table}[ht]
  \centering
  \small
  \caption{Pre-training hyperparameters for Objaverse and ShapeNet55-34 datasets.}
  \label{tab:hyperparams}
  \resizebox{\columnwidth}{!}{%
  \begin{tabular}{lcc}
    \toprule
    \textbf{Parameter} & \textbf{Objaverse 660K} & \textbf{ShapeNet55-34} \\
    \midrule
    Batch size & 8 & 8 \\
    Gradient accumulation steps & 8 & 8 \\
    Effective batch size & 64 & 64 \\
    Peak learning rate & 0.005 & 0.005 \\
    Weight decay & 0.05 & 0.05 \\
    Warmup epochs & 10 & 10 \\
    Total scheduled epochs & 300 & 300 \\
    Actual trained epochs & 24--35 & 71 \\
    AMP precision & bfloat16 & bfloat16 \\
    Gradient clipping max norm & 1.0 & 1.0 \\
    Center prediction weight ($\eta$) & 0.0--1.0 & 1.0 \\
    Input points & 8,192 & 8,192 \\
    Point features & 6 (xyz+rgb) & 3 (xyz only) \\
    \bottomrule
  \end{tabular}%
  }
\end{table}

\subsection{Pre-training Loss Curves}
\label{sec:loss_curves}

Pre-training was halted when the reconstruction loss entered a stable plateau. The final loss values for each variant at the end of pre-training are summarized in Table~\ref{tab:loss_curves}.

\begin{table}[ht]
  \centering
  \small
  \caption{Final loss values at pre-training termination.}
  \label{tab:loss_curves}
  \resizebox{\columnwidth}{!}{%
  \begin{tabular}{lccc}
    \toprule
    \textbf{Variant} & \textbf{Epochs} & \textbf{Final Chamfer Loss} & \textbf{Final Center Loss} \\
    \midrule
    Pre-V1 (PCP-MAE, MaskTransformer, Objaverse) & 24 & 0.57 & 1.41 \\
    V1 hybrid (PCP-MAE, MaskTransformer, Objaverse) & 24 & 0.57 & 1.41 \\
    V2 (PCP-MAE, PointTransformer, Objaverse) & 32 & 0.82 & 0.046 \\
    Point-MAE (Point-MAE, PointTransformer, Objaverse) & 35 & 0.93 & -- \\
    Mask Point-MAE (Point-MAE, MaskTransformer, Objaverse) & 31 & 0.97 & -- \\
    ShapeNet55-34 (PCP-MAE, MaskTransformer) & 71 & 0.84 & 4.25 \\
    \bottomrule
  \end{tabular}%
  }
\end{table}

Note: The center loss for ShapeNet55-34 remains higher than Objaverse variants because ShapeNet55-34 lacks color information, making center prediction more challenging due to less discriminative local features.

\end{document}